\documentclass[a4paper,twocolumn,fleqn]{article} 
\usepackage{ist}
\usepackage{amsmath}
\usepackage[american]{babel}
\usepackage{microtype}
\usepackage{graphicx}
\usepackage{algorithm}
\usepackage{algorithmic}
\usepackage{siunitx}
\usepackage{subcaption}
\usepackage{float}
\usepackage{pdfpages}
\usepackage{multirow}
\usepackage{mathtools}
\newcommand{\D}{\mathrm{d}}

\title{Designing a physically-feasible colour filter to make a camera more colorimetric }
\author{ Yuteng Zhu \\
School of Computing Sciences, University of East Anglia, Norwich, United Kingdom}
\date{}
\begin{document} 
\maketitle 
\thispagestyle{empty} 
%%%%%%%%%%%%%%%%%%%%%%%%%%%%%%%%%%
% Abstract
\begin{abstract}
Previously, a method has been developed to find the best colour filter for a given camera which  results in the new effective camera sensitivities  that best meet the Luther condition. That is, the new sensitivities are approximately linearly related to the XYZ colour matching functions. 
However, with no constraint, the filter derived from this Luther-condition based optimisation can be rather non-smooth and transmit very little light which are impractical for fabrication.

In this paper, we extend the Luther-condition filter optimisation method to allow us  to incorporate both the smoothness and transmittance bounds of the recovered filter which are key practical concerns. Experiments demonstrate that we can find physically realisable filters which are smooth and reasonably transmissive with which the effective `camera+filter' becomes significantly more colorimetric.

\end{abstract}

\section{Introduction} \label{sec:intro}

Digital cameras measure colours of an image with three sensors and use all three components (e.g.\ RGB values) to represent a pixel colour. If colour values produced by a camera can be linearly transformed to the ground-truth XYZ tristimulus values, we call the camera a colorimetric device~\cite{bibHorn}. Exact colour reproduction is ensured if the Luther condition is satisfied, i.e.\ the camera spectral sensitivities are  a linear transform from the CIE XYZ colour matching functions~\cite{bibLuther}. 
While most cameras can make reasonable colour measurements - their recorded RGBs, mapped by a $3\times$ 3 colour correction matrix are plausibly similar to the corresponding XYZ tristimuli - though they are far from being colorimetric. 

\begin{figure}[!ht]
    \centering
    \includegraphics[width=0.5\textwidth]{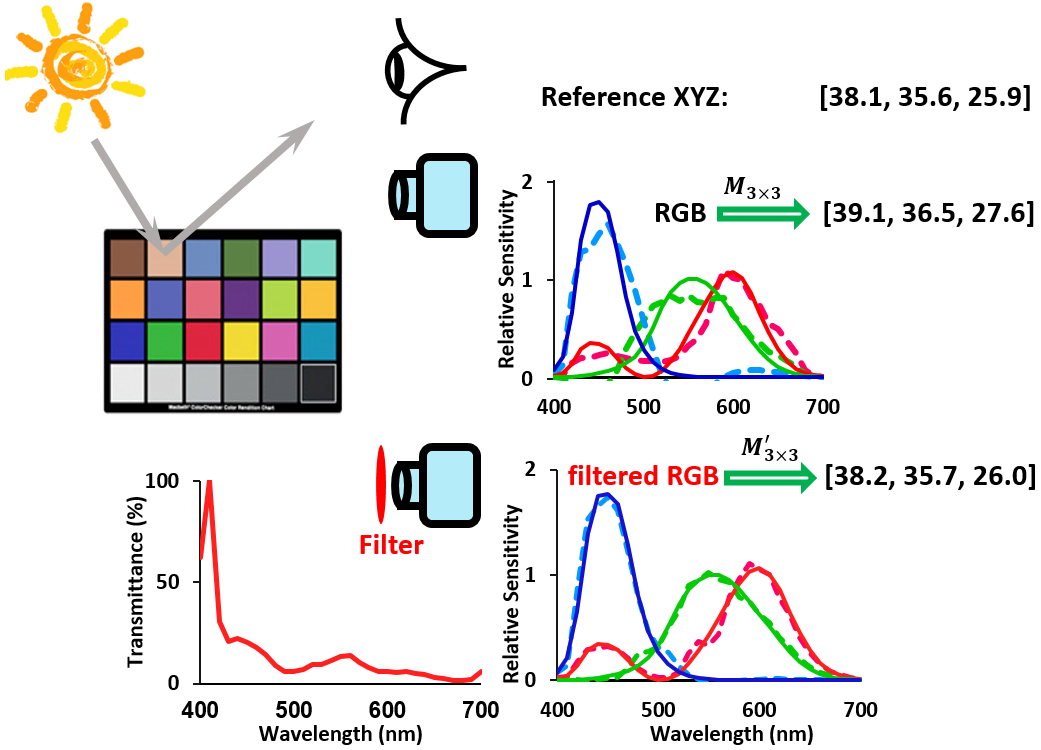}
    \vspace{0.1cm}
    \caption{An observer and a camera viewing the Macbeth Chart under D65. The linearly fitted camera spectral curves are plotted against the colour matching functions. By applying the optimised filter, the camera nearly meets the Luther condition with its responses better approximating to XYZ tristimulus values.}
    \label{fig:illustration}
    \vspace{0.5cm}
\end{figure}

Figure~\ref{fig:illustration} shows a human observer and a camera viewing the Macbeth ColorChecker under a standard CIE D65 illuminant~\cite{bibCIE}. The camera triplets (after the linear colour correction) and its XYZ values of the patch representing light skin colour are given. Even after the colour correction, the camera records rather different colour values compare to the  XYZs. Also shown are the colour sensitivities of the camera linearly fitted to the XYZ colour matching functions. Remember a camera is colorimetric only if the XYZ matching functions are similar to the fitted camera curves, here they are close but still quite different. 

Finlayson \emph{et al.}~\cite{bibLutherFilter} found a colour filter through optimisation such that the new effective `camera+filter' system returns the least error with respect to the Luther condition and thus makes the camera much more colorimetric. The filtered camera sensitivities linearly fitted to the XYZ colour matching functions are shown on the bottom of Fig.1. And, the resulting measured camera responses are now similar to the actual XYZs. Also the optimised filter is shown on the bottom left.

Clearly, from the figure the filter that renders the camera closest to satisfying the Luther condition has problems. For much of the visible spectrum it absorbs most of the incoming light. And, it varies sharply over the  spectrum which can mount great challenge to manufacture it precisely.
Potentially, the colorimetric performance of the filter will be vulnerably degraded to small perturbation in the fabrication process.   Thus the optimal filter obtained from the Luther-condition optimisation will not necessarily results in the optimal realisable filter.    

In this paper, we extend the Luther-condition based filter design~\cite{bibLutherFilter} to find a filter which is shaped in a smoothing manner with plausible transmissivity. We take inspiration from the work of 
Vora \emph{et al.}~\cite{bibVora} who emphasized the importance of the smoothness property of the filter transmittance curves for the fabrication process. 
This led Vora and Trussell~\cite{bibVora1997} proposed to use a sum-of-gaussian functions for constructing a smooth filter set for colour scanners. In our work smoothness will be enforced by specifying filters as a linear combination of the first few terms of a cosine basis expansion. Transmittance constraints are enforced by reformulating the minimisation as a linearly constrained quadratic optimisation problem~\cite{biblinearp}.

Experiments demonstrate that we can find highly transmissive  and smooth filters that can greatly improve the colorimetric performance of a camera and thus we can use the `camera+filter' setting for applications where high accuracy of colour measurement is needed. 

\section{Background}
\subsection{Image Formation}

The colour formation of a pixel, under the Lambertian surface model, can be written as 
\begin{equation}
    \underline{\rho} = \int_{\omega} \!E(\lambda) S(\lambda)\underline{Q}(\lambda)\,\D\lambda
    \label{eq:colorformation}
\end{equation}
where $E(\lambda)$, $S(\lambda)$ and $\underline{Q}(\lambda)$ respectively denote the spectral power distribution of the illuminant, the spectral reflectance function of the surface and a vector function of camera spectral sensitivities (i.e.\ R-, G-, and B- channels). 
The camera response $\underline{\rho}$ is a 3-component vector resulting from the product of these three factors integrating over the visible spectrum $\omega$. The product of $E(\lambda)S(\lambda)$ is termed as the \textit{colour signal}, $C(\lambda)$. Henceforth, we will use $C(\lambda)$ to represent the spectral product of the light and reflectance for convenience.

Practically, this continuous integration can be well approximated by the summation of the product of discretely sampled data:
\begin{equation}
    \underline{\rho} \approx \sum_{\lambda_i = \lambda_1}^{\lambda_n} C(\lambda_{i}) \underline{Q}(\lambda_{i})\Delta \lambda\\
    \label{eq:discrete}
\end{equation}
where the wavelength range from $\lambda_1$ to $\lambda_n$ are divided by the equal width $\Delta \lambda$~\cite{bibBibel}. In this paper we take the visible spectrum over which the summation in Eq.~(\ref{eq:discrete}) is made from 400 nm to 700 nm and using the commonly accepted sampling rate of $\Delta \lambda = 10$ nm therefore makes 31 sampling points for a spectrum. 

Equation 2 can be  written in the vector-matrix formula:
\begin{equation}
   \underline{\rho}=\underline{C}^T{Q}
    \label{eq:matrix}
\end{equation}
where $^T$ denotes the matrix transpose operator. $Q$ is an $31 \times 3$ matrix with each column denoting the sampled sensor spectral sensitivity curve (and the sampling distance is incorporated within $Q$). $\underline{C}$ is a 31-dimensional vector denoting a sampled colour signal spectrum. 

Let $X$ represent the sampled CIE1931 XYZ colour matching functions. The XYZ tristimulus response to a colour signal spectrum $\underline{C}$ is written as:
\begin{equation}
    \underline{x} = \underline{C}^T X
\end{equation}

\subsection{Luther Condition}
The Luther-condition is fully satisfied when  the spectral response curves of the camera become exactly a linear transform from the CIE XYZ colour matching functions. Mathematically, 
\begin{equation}
    X = QM
    \label{eq:Luther}
\end{equation}
where $M$ is a $3 \times 3$ mapping matrix.

If a camera meets the Luther-condition, for any two colour signals produced the same values by the camera sensors, they should also make the same XYZ stimulus values:
\begin{equation}
   \underline{C}^T_1 Q = \underline{C}^T_2 Q \; \xRightarrow{M} \; \underline{C}^T_1 X = \underline{C}^T_2 X
\end{equation}
and thus are indistinguishable to the observer. Here we gives a precis of the Luther condition. Readers are referred to ~\cite{bibHorn} for details.

\subsection{Luther-condition Based Filter Optimisation}

Let the vector $\underline{f}$ denotes the spectral transmittance of a colour filter. Placing a colour filter in front of a camera can be modeled as, at each sampled wavelength, the product of the filter and the camera response curves. This is mathematically written as $diag(\underline{f})Q$ where $diag()$ turns the filter vector into a diagonal matrix with its values on the main diagonal.

The Luther-condition based filter design proposed previously~\cite{bibLutherFilter} is to specify the spectral curve of a filter when placed in front of a given camera that best maps the new effective camera spectral sensitivities---after a linear correction---to the colour matching functions. The filter and the correction matrix that best satisfy such condition are found by minimising:
\begin{equation}
    	\min\limits_{\underline{f},M}\parallel{diag(\underline{f})QM - \text{\it X}}\parallel^2_F , \;\text{s.t.}\;\underline{f}>0
\label{eq:LutherOpt}
\end{equation}
where $\parallel \; \parallel^2_F$ denotes the square of the Frobenius norm (i.e.\ the sum of the squares of all elements in the matrix). Note that the filter is also constrained to be positive since a physical filter should have non-negative transmittance. This modified Luther-condition optimisation aims for the best $\underline{f}$ and $3 \times3 $ linear transform $M$ that return the least errors between the two spectral sensitivities sets.

\section{Luther Filter Optimisation under Constraints}

Let us denote the minimum and maximum filter transmittance bounds as:
\begin{equation}
    f_{min}\leq\underline{f}\leq f_{max}.
	\label{eq:bounds}
\end{equation}
Under such constraints, the filter transmittance should fall within the range of the upper threshold $ f_{max}$ and the lower threshold $ f_{min}$ over the whole spectrum.
For a physically realisable filter, it can transmit maximally 100\% of the incident light while in the extreme case, it can absorb all the lights (and therefore transmits 0\% light). Our desire is to let the filter overall transmit as much light as possible. 
This is achieved by raising up the lower boundary $f_{min}$ while keeping $f_{max} = $ 100\% (as fully transmissive).

We also would like to modify the filter design optimisation so that the solved-for filter is smooth. Let us represent  the $\underline{f}$ as belonging to a linear basis:
\begin{equation}
\underline{f}=B\underline{c} 
	\label{eq:smoothness}
\end{equation}
where $B$ has size of $31\times m$ where basis vectors are aligned up column-wisely and vector $\underline{c}$ contains the coefficients for each basis. 
In this paper, we adopt the discrete cosine series expansion~\cite{bibDCT} and take linear combinations of the first $m$ orthonormal basis to make smooth filters. Because the individual terms of the cosine basis are smooth, any linear combination of this basis is also smooth.

Now let us rewrite  the filter optimisation in Eq.~(\ref{eq:LutherOpt}) under constraints as
\begin{equation}
    	\min \limits_{\underline{c}, M} \parallel diag(B\underline{c})QM - {\it X} \parallel_F^2  \;\textrm{s.t.}\;
    	f_{min}\leq B\underline{c} \leq f_{max}
	\label{eq:pb_form_constraints}
\end{equation}

\subsection{Solving the Minimisation} \label{sec:ALS}

The optimisation in Eq.~(\ref{eq:pb_form_constraints}) has no closed-form solution. Fortunately, the coefficient vector $\underline{c}$ and matrix $M$ can be solved individually in closed form of least-squares regressions. 
Thus, we use the \textit{Alternating Least-Squares} (ALS) technique to solve for the optimisation.
Promisingly the alternating least-squares method is guaranteed to converge (although not necessarily to the global optimum)~\cite{bibConvergence}. 

Algorithm 1 presents the workflow of using the ALS method to find the filter and correction matrix. Specifically, we solve for the filter $\underline{f}$ by holding the matrix $M$ fixed (see step 4) and alternatively using the newly solved filter to solve for the matrix $M$ (see step 5) and the process will continue updating both matrices in turn until it converges to a predefined error threshold (see step 7). 

\begin{algorithm}[!hbt]
	\caption{ALS algorithm for the constrained Luther-condition filter optimisation}
	\begin{algorithmic}[1]
		\STATE{$i=0, \; Q^0 = Q$} \label{algo_init}
		\REPEAT
		\STATE{$i = i+1$}
     \STATE{$\min\limits_{\underline{c}^i} \parallel{diag(B\underline{c}^i)Q M^{i-1} -  X}\parallel_{F}^2,\; f_{min}\leq  B\underline{c}^i \leq f_{max}$} \label{codef}
  		\label{codeM2}
		\STATE{$\min\limits_{M^i} \parallel{diag(B\underline{c}^i) QM^{i} -  X}\parallel_{F}^2$}
		\label{codeM}		
		\STATE{$Q^{i}=  diag(B\underline{c}^{i})Q M^{i}$}
		\UNTIL{$\parallel{Q^{i} - Q^{i-1}}\parallel_{F}^2 \, < \, \epsilon $} \\
		\STATE{$ \underline{f} = \underline{f}^{i} $\quad and \quad $M = M^i $	}
	\end{algorithmic}
	\label{algo1}
\end{algorithm}

In the following, we will demonstrate how to in turn solve the filter and correction matrix in the least-squares way. Given a known coefficient vector $\underline{c}$ for the corresponding basis matrix $B$, the correction matrix $M$ can be solved by:
\begin{equation}
    M = [diag(B\underline{c})Q]^+X 
    \label{eq:M}
\end{equation}
where $^+$ denotes the Moore-Penrose pseudo-inverse~\cite{bibPseudoinverse}.

To solve for the filter, it is useful to {\it vectorize} the minimisation.
The vecterisation, denoted $vec()$, transforms a matrix to a vector by stacking its columns on top of one another. By using the vecterisation, we can rewrite the minimisation as:
\begin{equation}
    \parallel vec(diag(\underline{f})QM) - vec(X) \parallel_F^2
    \label{eq:vecOpt}
\end{equation}

Now let us rewrite the diagonal filter matrix as a summation of each value in the diagonal, $f_i$, with a single entry matrix $D_i$ as $diag(\underline{f}) = \sum_{i = 1}^{31} f_i D_i$. $D_i$ is a $31 \times 31$ matrix with a single non-zero entry  $D(i,i) = 1$. By substituting the filter matrix using this new representation into Eq.~(\ref{eq:vecOpt}), we obtain
\begin{equation}
 vec(diag(f)QM) = \sum_{i = 1}^{31} f_i \; vec(D_i  QM)
 \label{eq:vecsum}
\end{equation}
as the vectorisation operator and summation are commutative.

Denoting $V = [\underline{v}_1, \underline{v}_2, \cdots, \underline{v}_{31}]$ where $\underline{v}_i= vec(D_i  QM)$, Eq.~(\ref{eq:vecsum}) can be expressed more compactly using matrix-vector multiplication as
\begin{equation}
    vec(diag(f)QM) = V\underline{f}
\end{equation}
Note that $D_i  QM$ has the size of $31 \times 3$, so after vectorisation, $\underline{v}_i$ is a $93\times 1$ vector which makes $V$ a $93 \times 31$ matrix. 

Denoting $\underline{w}$ for $vec(X)$, the Luther-condition filter optimisation under constraints can be rewritten as 
\begin{equation}
    \min\limits_{\underline{c}}\parallel{VB\underline{c} -\underline{w}}\parallel^2_F\;\text{s.t.}\;f_{min}\leq  B\underline{c} \leq f_{max}
\end{equation}
The above minimisation can be 
solved as a quadratic problem subject to linear inequality constraints as
\begin{equation}
    \min \limits_{\underline{c}} \; \underline{c}^T B^TV^TVB \underline{c} - 2 \underline{w}^T VB \underline{c} \quad \text{s.t.}\;f_{min}\leq  B\underline{c} \leq f_{max}
\end{equation}

\section{Experiments and Results}
The Luther-condition filter optimisation is performed on a 
Canon 40D camera with measured spectral sensitivity functions~\cite{bibSensors}. In the first experiments, we are solving for the best filter of the testing camera using 6, 8 and 10 cosine basis functions. In these cases, we seek the filters that pass at least 20\% of the incident light. For each experimental condition, we solve for the best filter (equivalently the coefficients for the cosine basis) and the best correction matrix using Algorithm 1.

\begin{figure}[t]
	\begin{subfigure}{.45\textwidth}
		\includegraphics[width=\textwidth]{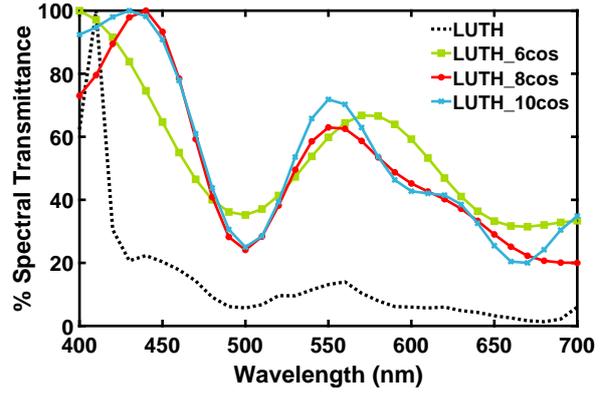}%
		\caption{under different smoothness constraints}
		\label{fig:20pct}
	\end{subfigure} 
	\vspace{0.8cm} \\
	\begin{subfigure}{.45\textwidth}
		\includegraphics[width=\textwidth]{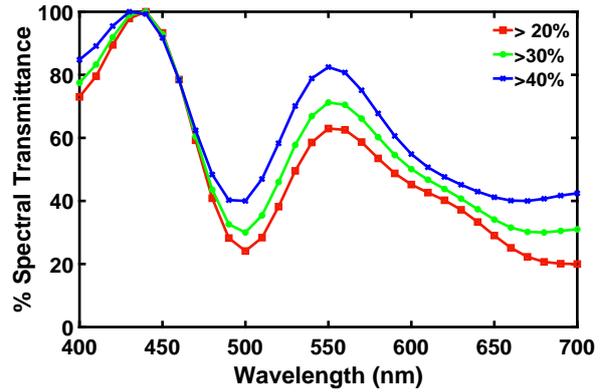}%
		\caption{under different transmittance constraints}
		\label{fig:8cos}
	\end{subfigure}
	\vspace{0.8cm}
	\caption{Spectral distribution of Luther-optimised filters. (a) filter with no constraint (dotted black line), filters constructed by 6-, 8- and 10-term cosine basis functions with the minimum transmittance of 20\% (coloured solid lines).  (b) filters with minimum transmittance of 20\%, 30\% and 40\% composed by 8-cosine basis.}
	\vspace{0.5cm}
\end{figure}

The  spectral transmittance distribution of the best filters under these conditions are shown in Fig.~\ref{fig:20pct}. For reference, we also plot the best Luther filter with no  constraint i.e.\ we run the original optimisation [5] (see the dashed black line denoted by `LUTH'). Evidently, the fewer cosine basis functions we use, the smoother the filter becomes.

To quantitatively measure the goodness of spectral sensitivities match between the camera sensors and XYZ colour matching functions (after linear transform), we calculate the normalised root-mean-square error (NRMSE). This error metric is defined as
\begin{equation}
    NRMSE = \frac{\parallel QM -X \parallel _F}{\parallel X \parallel _F}
\end{equation}
where $Q$ and $X$ are camera sensitivities and colour matching functions.  The best least-squares fit between two sets of spectral sensitivities can be found using Moore-Penrose inverse, i.e.\ $M = Q^+X = (Q^T Q)^{-1}QX$. 
Similarly, for the 'camera+filter' system, denoting ${\cal Q} = diag(\underline{f})Q$, the best mapping can be found as $M = {\cal Q}^+X = ({\cal Q}^T {\cal Q})^{-1}{\cal Q}X$. 

NRMSE gives a number between 0 and 1 meaning, respectively, not colorimetric at all and fully colorimtric.
The goodness of the spectral sensitivity fit in terms of NRMSE is given under the colum of {\bf NRMSE} in Table~\ref{tab:results} (see the first column). The best linear mapping from camera spectral sensitivities to CIE CMFs (no filter) is presented as a baseline reference in row 1 denoted \textbf{`Linear'} and together with the optimal Luther-filter with no particular constraint denoted \textbf{`LUTH'} in row 2. Results for Luther-condition filters under various smoothing (rows 3-5) constraints and transmittance boundedness (rows 6-8)  are also presented. 

From the column of {\bf NRMSE} in Table 1, it  can be seen that using Luther-condition optimised filters with constraints can still reduce the spectral mismatch error to half or even a third (when using 8- or 10- basis) comparing to the native linear correction (\textbf{`Linear'} ).  Encouragingly, the new filters - smoothing and relatively transmissive - are not far from the best performance results using the optimisation with no constraint.

\begin{table}[t]%
\centering
\vspace{0.2cm}
\renewcommand{\arraystretch}{1.5}
\setlength{\arrayrulewidth}{0.35mm}
\setlength\tabcolsep{4pt} 
\begin{tabular}{llcccc}
\hline                                      
& \multirow{2}{*}{\textbf{NRMSE}}   & \multicolumn{4}{c}{\textbf{CIELAB   error metrics}}  \\ \cline{3-6}
& & \textbf{Mean}     & \textbf{median}     & \textbf{95\%}   & \textbf{max}   \\\hline
\textbf{Linear} & 0.298    & 1.72              & 1.03                & 5.12              & 28.39       \\\hline
\textbf{LUTH}        & 0.062   & 0.44              & 0.22                & 1.48           & 8.76     \\\hline                 
& \multicolumn{5}{l}{\textbf{Filter transmittance $\underline{f} \geq$ 20\%  }} \\ \hline
\textbf{LUTH\_6cos}                                                       & 0.160                             & 0.94              & 0.54                & 2.84                & 21.14       \\\hline
\textbf{LUTH\_8cos}                                                       & 0.105                             & 0.62              & 0.38                & 2.01         & 9.53     \\\hline
\textbf{LUTH\_10cos}                                                      & 0.098                             & 0.69              & 0.42                & 2.27             & 10.06    \\\hline
& \multicolumn{5}{l}{\textbf{ Using an  8-term cosine basis}}    \\ \hline
\textbf{$\underline{f}\geq$20\%}                                           & 0.105                             & 0.62              & 0.38                & 2.01           & 9.53    \\\hline
\textbf{$\underline{f}\geq$30\%}                                           & 0.111                             & 0.69              & 0.41                & 2.22           & 12.69    \\\hline
\textbf{$\underline{f}\geq$40\%}           & 0.129                             & 0.83              & 0.46                & 2.63          & 16.62    \\\hline
\end{tabular}
\vspace{0.2cm}
\caption{Goodness of sensitivities match error (NRMSE) and colour measurement results ($\Delta E^{*}_{ab}$ Statistics) of using {\bf Linear} correction only (without a filter), using linear correction and the {\bf LUTH}er-condition optimised filter (without constraint), and using linear correction and {\bf LUTH}er-condition optimised filters under various constrained conditions.}
\label{tab:results}
\vspace{0.5cm}
\end{table}%

We also evaluate how good the filtered camera performs in the colour measurement experiments in terms of perceptual colour errors -- CIELAB colour difference metrics~\cite{bibBibel}. The colour correction experiments are performed for a set of 102 illuminants and 1995 reflectance spectra~\cite{bibSFU}. We calculate the RGBs of all reflectance spectra under each illuminant, and find the best $3 \times 3$ correction matrices mapping RGBs to the ground-truth XYZs before converted into CIELAB colour space. Then the overall mean, median, 95-percentile and max of $\Delta E_{ab}^*$ are averaged over the all test lights. 

The colour measurement results are given in Table 1 (see the columns under CIELAB error metrics). Under the 20\% transmittance bound, we can see that the optimal filter  linearly composed by 8 cosine basis (see the red line in Fig.~\ref{fig:20pct} denoted \textbf{`LUTH\_8cos'}) outperforms the other two smoothness conditions: it reduces nearly two-thirds of the colour errors across all statistical metrics comparing to those by the linear colour correction. Although constrained filters perform less well than the non-constrained Luther optimal filter, it gains greatly on the filter transmissivity and smoothness.  

The best filter for minimum transmittances of 20\%, 30\% and 40\% are also calculated.
The spectral distribution of the solved-for filters are shown in Fig.~\ref{fig:8cos} and their colour correction results are given in the bottom rows in Table 1. As expected, greater minimum threshold will lead to less effective colour reduction. Yet a filter having a minimum transmittance value of 40\% (an overall of 63\% transmittance which is reasonably transmissive) can still make a camera significantly more colorimetric by reducing nearly half of the colour errors. 

\section{Conclusion}
In this paper we extend the Luther-condition filter optimisation by providing a method for incorporating  smoothness and transmittance bounds into the same optimisation framework. We find that the solved filters which are smooth and reasonably highly transmissive - and so plausibly manufacturable - provide similar performance to the prior art of unconstrained minimisation (which produces a non-smooth filter that transmits almost no light).

%%%%%%%%%%%%%%%%%%%%%%%%%%%%%%%%%%
% Bibliography

\begin{biography}
Yuteng Zhu is currently a PhD candidate under the supervision of Prof. Finlayson in the Colour and Imaging Laboratory at the University of East Anglia, UK. She received her dual MSc degree from Zhejiang University, China, and \'Ecole Centrale de Marseille, France, in 2014. She is working on filter design for accurate colour measurement by digital cameras. \end{biography}

\end{document}